\makeatletter
\@ifundefined{isChecklistMainFile}{
	\newif\ifreproStandalone
	\reproStandalonetrue
}{
	\newif\ifreproStandalone
	\reproStandalonefalse
}
\makeatother

\ifreproStandalone

\documentclass[letterpaper]{article} 
\usepackage{aaai2026}  
\usepackage{times}  
\usepackage{helvet}  
\usepackage{courier}  
\usepackage[hyphens]{url}  
\usepackage{graphicx} 
\usepackage{natbib}  
\usepackage{caption} 
\usepackage{newfloat}
\usepackage{listings}
\DeclareCaptionStyle{ruled}{labelfont=normalfont,labelsep=colon,strut=off} 
\usepackage[ruled, vlined, linesnumbered, lined, boxed, commentsnumbered]{algorithm2e}
\SetKwRepeat{Do}{do}{while}
\usepackage{amsthm}
\usepackage{amsmath} 
\usepackage{multirow}
\usepackage{makecell}
\usepackage{amsfonts}
\usepackage{subcaption}

\newcommand{\rational}{\mathbb{R}}
\newcommand{\integer}{\mathbb{Z}}
\newcommand{\ithRow}{A_{i \bullet}}
\newcommand{\kthRow}{A_{k \bullet}}
\newcommand{\xthRow}[1]{A_{ #1 \bullet}}

\newcommand{\set}[1]{\{ #1 \}}

\newcommand{\false}{\bot}
\newcommand{\conflict}{\Phi_{\false}}
\newcommand{\sysLinCons}{\Phi}
\newcommand{\assign}{\leftarrow}

\newcommand{\size}[1][]{|#1|}

\newcommand{\sharpP}{{\sf \# P}}

\DeclareMathOperator*{\norm}{norm}

\newcommand\ie{{\it i.e.}}
\newcommand\eg{{\it e.g.}}
\newcommand\aka{{\it a.k.a. }}

\newtheorem{definition}{Definition}
\newtheorem{example}{Example}

\nocopyright

\title{An Exhaustive DPLL Approach to Model Counting over \\ Integer Linear Constraints with Simplification Techniques}
\author {
    Mingwei Zhang\textsuperscript{\rm 1},
    Zhenhao Gu\textsuperscript{\rm 1},
    Liangda Fang\textsuperscript{\rm 1},
    Cunjing Ge\textsuperscript{\rm 2},
    Ziliang Chen\textsuperscript{\rm 3},
    Zhao-Rong Lai\textsuperscript{\rm 1},
    Quanlong Guan\textsuperscript{\rm 1}
}
\affiliations {
    \textsuperscript{\rm 1}College of Information Science and Technology, Jinan University\\
    \textsuperscript{\rm 2}AI School, Nanjing University\\
    \textsuperscript{\rm 3}Pengcheng Laboratory\\
    mingweizhang@stu2022.jnu.edu.cn
}

\usepackage{bibentry}

\begin{document}

\maketitle

\begin{abstract}
	\looseness=-1
Linear constraints are one of the most fundamental constraints in fields such as computer science, operations research and optimization.
Many applications reduce to the task of model counting over integer linear constraints (MCILC).
In this paper, we design an exact approach to MCILC based on an exhaustive DPLL architecture.
To improve the efficiency, we integrate several effective simplification techniques from mixed integer programming into the architecture.
We compare our approach to state-of-the-art MCILC counters and propositional model counters on 2840 random and 4131 application benchmarks.
Experimental results show that our approach significantly outperforms all exact methods in random benchmarks solving 1718 instances while the state-of-the-art approach only computes 1470 instances.
In addition, our approach is the only approach to solve all 4131 application instances.
\end{abstract}


\section{Introduction}
\looseness=-1
Linear constraint is one of the most fundamental constraints in fields of computer science \cite{UlrNW2023}, operations research \cite{Shv2025} and optimization \cite{KorP2024}.
Given a set of integer linear constraints (that is, linear constraints with only integer variables), a solution (\aka model) is an integer vector satisfying all of the constraints.
Many applications reduce to the model counting problem of integer linear constraints (MCILC), for example, counting-based search \cite{ZanP2009,Pes16}, temporal planning \cite{HuaLO2018}, program analysis \cite{GelDV2012,LucPD2014}, and combinatorics \cite{DesZ2020,HarRK2021,GamaK2010}.

\looseness=-1
MCILC is a computationally demanding task and was proved to $\sharpP$-complete \cite{Les1979}.
Numerous approaches to MCILC fall into two categories: approximate and exact.
Among approximate approaches, \citet{GeMMZHZ2019} estimates the number of solutions over integer linear constraints via computing the volume of feasible rational region defined by linear constraints which is solved by existing tools: \texttt{PolyVest} \cite{GeM2015} and \texttt{Vinci} \cite{BueEF2000}.
Recently, \citet{Ge2024} adopted hit-and-run random walk sampling method \cite{Lov1999} to compute an approximate estimate for MCILC.

\looseness=-1
In contrast, exact approaches have also been developed, aiming to offer the precise number of solutions.
\citet{Bar1994a} designed a seminal exact algorithm for MCILC.
As a set of linear constraints can be visualized as a polytope, MCILC can be reduced to the problem of calculating the number of integer points within this polytope.
\citeauthor{Bar1994a}'s algorithm decomposes a polytope into a short signed sum of unimodular cones, each of which admits an explicit generating function.
Each generating function is expressed as a power series where each monomial corresponds to an integer point in the associated unimodular cone.
\texttt{LattE} \cite{LoeHT2004} and its successor \texttt{Barvinok} \cite{VerSB2007} are the implementations of \citeauthor{Bar1994a}'s algorithm.
Unfortunately, \citeauthor{Bar1994a}'s algorithm does not scale well to instances with more than 15 variables.
\citet{GeB2021} introduced a preprocessing strategy that factorizes a MCILC problem into multiple smaller subproblems via eliminating rows and columns from the original set of linear constraints.
Each subproblem is then solved by \citeauthor{Bar1994a}'s algorithm.
An alternative approach first translates a set of integer linear constraint into a propositional formula \cite{AbiS2014}, and then applies well-established exact model counters for propositional logic, such as \texttt{Cachet} \cite{SangBB2004}, \texttt{SharpSAT} \cite{Thu2006}, \texttt{D4} \cite{LagM2017b}, \texttt{ExactMC} \cite{LaiMY2021}, and \texttt{SharpSAT-TD} \cite{KorJ2021}.
In particular, \texttt{SharpSAT-TD} is able to solve more MCILC instances.

\looseness=-1
The aforementioned propositional model counters employs the exhaustive DPLL architecture \cite{BirL99} integrated with connected component-based decomposition \cite{JrP2000} as their main algorithm.
\citet{KorLM2015} extended this exhaustive DPLL architecture to finite-domain constraint networks and designed the algorithm \texttt{Cn2mddg} for compiling such networks into multivalued decomposable decision graphs (MDDGs).
Since MDDGs support tractable model counting and integer linear constraints are a special case of constraint networks, \texttt{Cn2mddg} offers an effective approach to MCILC.

\looseness=-1
However, \texttt{Cn2mddg} suffers from two key limitations when applied to MCILC.
(1) It compiles the entire constraint network into a MDDG, incurring a substantial space overhead. 
(2) As a general-purpose compiler for constraint networks, it is not optimized specifically for integer linear constraints.
In particular, it does not take advantage of some effective simplifications \cite{Sav1994,AndA1995,FugM2005,AchBG2020,GleHG2023} tailored for mixed integer programming that is able to accelerate MCILC solving.
Hence, \texttt{Cn2mddg} performs poorly on MCILC, demonstrated in our experimental results.

\looseness=-1
This paper is intended to design an exact model counter tailored for integer linear constraints. 
The main algorithm of our model counter is the exhaustive DPLL architecture enhanced with connected component-based decomposition. 
Furthermore, we incorporate several simplification techniques commonly used in mixed integer programming (including removing variables, strengthening bounds, strengthening coefficients, removing rows).
We evaluate our approach against state-of-the-art MCILC counters and propositional model counters on 2840 random and 4131 application benchmarks.
Experiment results show that our approach significantly outperforms all exact methods in random benchmarks solving 1718 instances in comparison to 1470, which is the most solved by the propositional model counter \texttt{SharpSAT-TD} augmented with the preprocessor \texttt{Arjun}.
In addition, our approach is the only approach to solve all 4131 application instances.

\section{Preliminaries}
\looseness=-1
In this section, we introduce the concepts of integer linear constraints and provide the definition of model counting over integer linear constraints.

\begin{definition} \rm
    The system of integer linear constraints $\sysLinCons$ is defined as a tuple $(A, b, l, u, M, N)$ where
    \begin{itemize}
        \item $A$: a $m \times n$ real matrix $\rational^{m \times n}$;
        \item $b$: a $m$-dimensional vector $\rational^{m}$;
        \item $l$: a $n$-dimensional vector $\integer^n$ denoting the lower bound of each variable;
        \item $u$: a $n$-dimensional vector $\integer^n$ denoting the upper bound of each variable;
        \item $M$: a set of positive integers where each integer denotes a row index of $A$;
        \item $N$: a set of positive integers where each integer denotes a column index of $A$.
    \end{itemize}
\end{definition}

\looseness=-1
We use $a_{ij}$ for the entry of the matrix $A$ with the $i$-th row and $j$-th column and $\ithRow$ for the $i$-th row vector of $A$.
We use $N_{-j}$ for the set $\set{k \in N \mid k \neq j}$ and $N_z(\ithRow) = \set{j \in N \mid a_{ij} \neq 0}$ for the set of indices of variables with non-zero coefficients in $\ithRow$.
We use $\size[M]$ and $\size[N]$ for the size of the two sets $M$ and $N$, respectively.
We use $x_i$ for the $i$-th entry of the unknown vector $x$.

\begin{example}
	\label{exa:LC}
	Consider the following system of constraints $\sysLinCons = (A, b, l, u, M, N)$ where $A = \begin{bmatrix}
		1  & -1 & 1  \\
		1  & 2  & 1  \\
		-1 & 1  & 3  \\
		-2 & -1 & -3
	\end{bmatrix}$,
	$b = \begin{bmatrix}
		1 \\
		3 \\
		2 \\
		4
	\end{bmatrix}$,
	$l = \begin{bmatrix}
		0 \\
		0 \\
		0 
	\end{bmatrix}$,
	$u = \begin{bmatrix}
		3 \\
		3 \\
		3 
	\end{bmatrix}$,
	$M = \set{1, 2, 3, 4}$, and $N = \set{1, 2, 3}$.
	
	The system $\sysLinCons$ corresponds to $4$ linear constraints with $3$ variables $x_1$, $x_2$ and $x_3$ defined as
	\begin{center}
		$\begin{cases}
			x_1 - x_2 + x_3 \leq 1    &  \\
			x_1 + 2x_2 + x_3 \leq 3   &  \\
			-x_1 + x_2 + 3x_3 \leq 2  &  \\
			-2x_1 - x_2 - 3x_3 \leq 4 & 
		\end{cases}$
	\end{center}
	where $x_1,x_2,x_3 \in [0, 3]$. \qed	
\end{example}

\looseness=-1
Given a system of linear constraints $\sysLinCons = (A, b, l, u, M, N)$, if $M = \emptyset$, then no linear constraint needs to be satisfied and hence $\sysLinCons$ is \textit{valid}.
In this case, the number of solutions is $\prod_{j \in N} (u_j - l_j + 1)$.   
We use $\conflict = (A_\false, b_\false, l_\false, u_\false, M_\false, N_\false)$ for the \textit{inconsistent} system of linear constraints where $A_\false = \begin{bmatrix} 0 \end{bmatrix}$, $b_\false = \begin{bmatrix} -1 \end{bmatrix}$, $l_\false = u_\false = \begin{bmatrix} 0 \end{bmatrix}$, $M_\false = \set{ 1}$, and $N_\false = \set{1}$, corresponding to the linear constraint $0 \cdot x_1 \leq -1$.

\looseness=-1
Any system of constraints with operators $\geq$, $<$, $>$ and $=$ can be rewritten into an equivalent system with only operator $\leq$.
For example, $a_{1} x_1 + \cdots + a_n x_n \geq b$ is equivalent to $(-a_1) x_1 + \cdots + (-a_n) x_n \leq -b$.
Therefore, we assume that each constraint in this paper involves only the operator $\leq$ for simplicity.

\looseness=-1
Let $M' \subseteq M$ and $N' \subseteq N$.
$A_{i N'}$ represents the vector $\ithRow$ restricted to the row indices in $N'$, and $A_{M'N'}$ denotes the matrix $A$ restricted to the row indices in $M'$ and the column indices in $N'$.
Likewise, we use $x_{N'}$ to denote the vector $x$ restricted to $N'$.


\looseness=-1
The \textit{maximal activity} of $\ithRow$ is defined as $\sup(\ithRow) = \sum_{j \in N, a_{ij} > 0} a_{ij}u_j + \sum_{j \in N, a_{ij} < 0} a_{ij}l_j$, and the \textit{minimal activity} as $\inf(\ithRow) = \sum_{j \in N, a_{ij} > 0} a_{ij}l_j + \sum_{j \in N, a_{ij} < 0} a_{ij}u_j$.
Similarly, the maximal and minimal activity of $\ithRow$ on the subset $N' \subseteq N$ can be defined as $\sup(A_{iN'})$ and $\inf(A_{iN'})$, respectively.

\begin{example}
    Consider the 3rd row of $A$, with $0 \leq x_1,x_2,x_3 \leq 3$, mentioned in Example~\ref{exa:LC}.
    The maximal activity of $A_{3 \bullet}$ is $\sup(A_{3 \bullet}) = -1 \times 0 + 1 \times 3 + 3 \times 3 = 12$,
    while the minimal activity $\inf(A_{3 \bullet})$ is $-3$.
    Similarly, for $N_{-2} = \set{1, 3}$, the maximal activity $\sup(A_{3N_{-2}}) = -1 \times 0 + 3 \times 3 = 9$,
    and the minimal activity $\inf(A_{3N_{-2}})$ is $-3$. \qed
\end{example}

\looseness=-1
A solution (\aka model) for a system of integer linear constraints $\sysLinCons$ is an integer vector $x$ s.t. $Ax \leq b$ and $l \leq x \leq u$.
The model counting problem for $\sysLinCons$ is to compute the number of solutions to $\sysLinCons$, denoted by $\#(\sysLinCons)$.

\begin{example}
	The solutions to $\sysLinCons$, defined in Example~\ref{exa:LC}, contain $\begin{bmatrix}0, 0, 0\end{bmatrix}^{\top}, \begin{bmatrix}0, 1, 0\end{bmatrix}^{\top}, \begin{bmatrix}1, 0, 0\end{bmatrix}^{\top}, \begin{bmatrix}1, 0, 1\end{bmatrix}^{\top}, \begin{bmatrix}1, 1, 0\end{bmatrix}^{\top}, \\  \begin{bmatrix}2, 0, 0\end{bmatrix}^{\top}, \begin{bmatrix}2, 0, 1\end{bmatrix}^{\top}, \begin{bmatrix}3, 0, 0\end{bmatrix}^{\top}$. Thus $\#(\sysLinCons) = 8$. \qed
\end{example}

\section{An Exhaustive DPLL Approach}
\looseness=-1
In this section, we propose an exhaustive DPLL approach to model counting over integer linear constraints. 
We first illustrate the main algorithm of our proposed approach (Sec. 3.1), and then introduce two key components: selecting variables (Sec. 3.2) and simplification techniques (Sec. 3.3).

\subsection{The Main Algorithm}
\begin{algorithm}[t!]
	\caption{{EDPLLSim}}
	\label{alg:main}
	\KwIn{$\sysLinCons = (A, b, l, u, M, N)$: A system of integer linear constraints;}
	\KwOut{$n$: The number of solutions to $\sysLinCons$.}
	
	\lIf{{\tt Cache}$(\sysLinCons) \neq nil$}
	{\Return {\tt Cache}$(\sysLinCons)$}
	
	$\sysLinCons \assign$ {\tt Simplify}$(\sysLinCons)$
	
	\lIf{$M = \emptyset$}
	{
		\Return $\prod_{j \in N} (u_j - l_j + 1)$
	}
	\lElseIf{$\sysLinCons = \conflict$}
	{
		\Return{$0$}
	}
	
	\If{$\sysLinCons$ is decomposable}
	{		
		$\Sigma \assign$ {\tt Decompose}$(\sysLinCons)$
		
		$n \assign 1$ 
		
		
		\ForEach{$\sysLinCons' \in \Sigma$}
		{
			$n \assign n \times$ {\tt EDPLLSim}$(\sysLinCons')$
		}
%
	}
	\Else
	{		
		$x_j \assign ${\tt SelectVariable}$(\sysLinCons)$
		
		$n  \assign 0$
		
		
		\ForEach{$v \in [l_j, u_j]$} 
		{
			$n \assign n + ${\tt EDPLLSim}$(\sysLinCons_{[x_j = v]})$
		}
	}
	
	{\tt Cache}$(\sysLinCons) \assign n$
	
	\Return{$n$}
	
\end{algorithm}

\looseness=-1
Algorithm \ref{alg:main} provides the pseudo-code of the main algorithm that performs exhaustive DPLL search over candidate solutions of integer linear constraints.
It takes a system $\sysLinCons$ of integer linear constraints as input, and outputs the number $\#(\sysLinCons)$ of solutions to $\sysLinCons$.
It first check whether the system is encountered before during the entire computation (line 1).
If so, we simply return {\tt Cache}$(\sysLinCons)$ to avoid duplicate computation.
Then the system $\sysLinCons$ will be simplified by some existing simplification techniques (line 2).
These simplification techniques transform the system $\sysLinCons$ into an equivalent one via removing redundant columns and rows from $\sysLinCons$ and strengthening coefficients and bounds in $\sysLinCons$ with the aim of accelerating the subsequent computation process.
The details of each presolving technique will be elaborated in Section 3.3.
After simplification, $\sysLinCons$ may become valid or inconsistent (lines 3 - 4).
In this case, it directly returns $\prod_{j \in N} (u_j - l_j + 1)$ or $0$.

\looseness=-1
After simplification, $\sysLinCons$ will be decomposed into a set of subsystems in two ways.
The first way obtains a partition $\Sigma: \set{\sysLinCons_1, \cdots, \sysLinCons_d}$ of the system $\sysLinCons$ where $\sysLinCons_i = (A, b, l, u, M_i, N_i)$ for each $1 \leq i \leq d$ (lines 6 - 9).
Moreover, it satisfies the three requirements: (1) $N_i \cap N_j = \emptyset$ for $i \neq j$, (2) $\bigcup_{i = 1}^d N_i = N$, (3) $M_i : \set{m \in M \mid \exists n \in N_i.a_{mn} \neq 0}$ and (4) $\#(\sysLinCons) = \prod_{i = 1}^{d}\#(\Sigma_i)$.
One simple method to obtain the desired partition is based on connected components of the primal graph of $\sysLinCons$.
The primal graph $G = (V, E)$ of $\sysLinCons$ is defined as $V = N$ and $(j, k) \in E$ iff there is a row index $i \in M$ s.t. $j,k \in N_z(\ithRow)$ with $j \neq k$.
All connected components of a graph $G$ can be generated in linear time in the size of $G$ using depth-first or breath-first search \cite{HopT1973}.
The second way first selects a variable $x_j$ of $\sysLinCons$ and then splits $\sysLinCons$ into $\Sigma = \set{\sysLinCons_{[x_j = v]} \mid l_j \leq v \leq u_j}$ by enumerating the value of $x_j$ (lines 11 - 14).
The subsystem $\sysLinCons_{[x_j = v]}$ is defined as $(A, b', l, u, M, N_{-j})$ where $b'_i = b_i - a_{ij} v$.
It can be easily verified that $\#(\sysLinCons) = \sum_{v = l_j}^{u_j}\#(\sysLinCons_{[x_j = v]})$.
The first decomposition way is our priority because of the following reason.
For simplicity, we assume that all variables have the same lower bound $l$ and upper bound $u$, and that $\sysLinCons$ can be decomposed into two subsystems $\sysLinCons_1$ and $\sysLinCons_2$ with disjoint sets $N_1$ and $N_2$ of variables.
The model counting time for $\sysLinCons$, denoted by $T(\sysLinCons)$, is highly sensitive to the number of variables in $\sysLinCons$.
In the worst case, it is necessary to exhaustively enumerate all possible valuations on each variable so as to compute the number of solutions to $\sysLinCons$ (more formally, $T(\sysLinCons) \in O((u - l)^{\size[N]})$ where $\size[N]$ denotes the size of $N$.
By the first decomposition way, the counting time $T(\sysLinCons)$ for $\sysLinCons$ is approximately $T(\sysLinCons_1) + T(\sysLinCons_2) \in O((u - l)^{\size[N_1]} + (u - l)^{\size[N_2]})$.
In contrast, $T(\sysLinCons) = \sum_{v \in [l, u]} T(\sysLinCons_{[x_j = v]}) \in O((u - l)^{\size[N]})$ in the second composition way.
Clearly, the first way is more efficient than the second one.

\looseness=-1
Finally, the result is stored in the cache and is returned (lines 15 - 16).
The main algorithm is guaranteed to terminate since the number of variables strictly decreases in two decomposition ways.
\subsection{Selecting Variable}

\looseness=-1
As discussed in Section 3.1, the first decomposition way is the preferred approach to decomposing a system of linear constraints.
Suppose that $\sysLinCons$ can be decomposed into two subsystems $\sysLinCons_1$ and $\sysLinCons_2$ with disjoint sets $N_1$ and $N_2$ of variables.
The counting time $T(\sysLinCons)$ for $\sysLinCons$ drops considerably when the size of sets $N_1$ and $N_2$ of variables for two subsystems $\sysLinCons_1$ and $\sysLinCons_2$ are balanced.
This is because $(u - l)^{\size[N_1]} + (u - l)^{\size[N_2]}$ reaches the minimum value when $\size[N_1] = \size[N_2] - 1$, $\size[N_1] = \size[N_2]$, or $\size[N_1] = \size[N_2] + 1$.

\looseness=-1
As mentioned in \cite{KorLM2015}, \textit{betweenness centrality} \cite{Bra2001,Bra2008} is a good heuristic for selecting variables in the second decomposition way.
It is beneficial for generating two components with balanced number of variables for each subsystem $\sysLinCons_{[x_j = v]}$.
Therefore, we adopt it to select a variable heuristically in line 11 in Algorithm \ref{alg:main}.
Betweenness centrality is a measure of the centrality of a node in a graph.
Given the primal graph $G = (V, E)$ of $\sysLinCons = (A, b, l, u, M, N)$, the score $bc(j)$ of the variable $x_j$ is defined as 
$$\sum_{k \in V \setminus \set{j}} \sum_{l \in V \setminus \set{j, k}} \frac{\sigma_{l}(j, k)}{\sigma(j, k)}$$ 
where $\sigma(j, k)$ denotes the number of the shortest paths from $j$ to $k$ and $\sigma_{l}(j, k)$ represents the number of those paths passing through $l$.
The betweenness centrality scores of all variables can be computed in $O(\size[V] \size[E])$.



\subsection{Simplification Techniques}
\label{sec:simplify}
\looseness=-1
We discuss several simplification techniques for integer linear constraints proposed in mixed integer programming.



\looseness=-1
\subsubsection{Removing Variables}
For each variable $x_j$, it will check if the lower bound $l_j$ equals to the upper bound $u_j$.
If so, we directly obtain a new system $\sysLinCons_{[x_j = u_j]}$ which denotes a system that is obtained from $\sysLinCons$ by removing the variable $x_j$.
Removing variables can be done in $O(\size[N])$. 

\looseness=-1
\subsubsection{Strengthening Bounds}
This simplification process employs the method of constraint propagation \cite{Sav1994,FugM2005}, which attempts to obtain tighter lower and upper bounds of each variable.
It traverses each linear constraint and strengthens both bounds on each variable with a non-zero coefficient.
The $i$-th constraint can be written as $A_{i N_{-j}}x_{N_{-j}} + a_{ij}x_j \leq b_i$.
If $a_{ij} > 0$, then the inequality $x_j \leq \frac{b_i - A_{i N_{-j}} x_{N_{-j}}}{a_{ij}} \leq \frac{b_i - \inf(A_{iN_{-j}})}{a_{ij}}$ holds.
Thus, the upper bound $u_j$ is updated to $\lfloor \frac{b_i - \inf(A_{iN_{-j}})}{a_{ij}} \rfloor$ when the current upper bound is greater than $\lfloor \frac{b_i - \inf(A_{iN_{-j}})}{a_{ij}} \rfloor$. 
Likewise, when $a_{ij} < 0$, the lower bound $l_j$ is replaced by $\lceil \frac{b_i - \inf(A_{iN_{-j}})}{a_{ij}} \rceil$ when $\lceil \frac{b_i - \inf(A_{iN_{-j}})}{a_{ij}} \rceil > l_j$.
It can be observed that the domain of each variable may be narrowed after performing strengthening bound.
Then, it is possible that the lower bound $l_i$ is greater than the upper bound $u_i$ for some variables $i$.
In this case, we mark $\sysLinCons$ as inconsistent.
In order to compute each $\inf(A_{i N_{-j}})$ for $j \in N$, we first compute $\inf(\ithRow)$ and then subtracting $a_{ij} u_j$ (resp. $a_{ij} l_j)$ from $\inf(\ithRow)$ individually when $a_{ij} < 0$ (resp. $a_{ij} \geq 0$).
In summary, the time complexity of strengthening bound is $O(\size[M] \size[N])$.


\looseness=-1
\subsubsection{Strengthening Coefficients}
\citet{Sav1994} proposed a technique, namely \textit{strengthening coefficients}, with the aim of obtain tighter coefficients of integer linear constraints.
Linear programming can be accelerated when applied to such strengthened constraints.
Accordingly, simplification techniques that rely on linear programming also benefits from integer linear constraints with tighter coefficients.
It traverses each constraint $i \in M$ and each variable $j \in N$.
For $a_{ij} > 0$, we let $d = b_i - \sup(A_{i N_{-j}}) - a_{ij} (u_j - 1)$.
If $a_{ij} \geq d > 0$ and $x_j \leq u_j - 1$, we replace $i$-th constraint by $A_{i N_{-j}} x_{N_{-j}} + (a_{ij} - d)x_j \leq b_j - du_j$ since they are equivalent on solutions and the new one has tighter coefficient for $x_j$.
Likewise, for $a_{ij} < 0$, we let $d = b_i - \sup\{A_{i N_{-j}}\} - a_{ij}(l_j + 1)$.
If $-a_{ij} \geq d > 0$, then the $i$-th constraint can be replaced with $A_{i N_{-j}} x_{N_{-j}} + (a_{ij} + d) x_j \leq b_i + d l_j$.
The time complexity of this technique is $O(\size[M] \size[N])$.


\looseness=-1
\subsubsection{Removing Rows} We hereafter introduce four methods to removing rows.
The first method, namely \textit{removing individual rows}, traverses each constraint.
If $\inf(\ithRow) > b_i$, then it means the $i$-th constraint cannot be satisfied, and thus we directly returns an inconsistent system $\conflict$.
If $\sup(\ithRow) \leq b_i$, then the $i$-th constraint is satisfied in every possible valuations on each variable, and thus we remove this constraint.
This first approach can be done in $O(\size[M] \size[N])$.

\looseness=-1
\citet{Sav1994} improved the first method by computing lower and upper bounds of each $A_i x$ via linear programming instead of using the two notions $\inf(\ithRow)$ and $\sup(\ithRow)$, respectively, which we call \textit{removing individual rows with linear programming}.
Given a system $\sysLinCons = (A, b, l, u, M, N)$, we use the modified system $\sysLinCons = (A, b, l, u, M \setminus \set{i}, N)$ as the constraints and $A_i x$ as the objective function.
Linear programming generates a valuation on $x$ s.t. $A_i x$ reaches the upper bound.
Similarly, if we use $- A_i x$ as the objective function, then we obtain the lower bound of $A_i x$.
The complexity of the second method is $O(\size[M] \size[N] (\size[M] + \size[N])^{1.5} L)$ where $L$ is bounded by the number of bits in the input since it invokes linear programming solver $\size[M]$ times and linear programming can be solved in $O(\size[N] (\size[M] + \size[N])^{1.5} L)$ in the worst case \cite{Vai1989}.



\looseness=-1
The third method, namely \textit{removing parallel rows}, aims to identify parallel constraints in $\sysLinCons$ \cite{AndA1995}.
We say the $i$-th constraint is \textit{parallel} to the $k$-th constraint iff there is a ratio $t$ such that $\ithRow = t \xthRow{k}$.
Suppose that the $i$-th and $k$-th constraint are parallel relative to a ratio $t$.
If $t > 0$ and $b_i < tb_k$, then the $k$-th constraint is redundant and thus being removed.
If $t < 0$ and $b_i < tb_{k}$, then both constraints cannot be satisfied simultaneously and thus this system is inconsistent.
This third method can be done in $O(\size[M]^2\size[N])$ since it pairwise compares every two constraints in $\sysLinCons$.
However, in the case where $A$ is an integer matrix, the time complexity can be reduced to $O(\size[M] (\size[N] + \log(\max(A))))$ where $\max(A)$ is the maximum value of entries in the matrix $A$ by introducing a hash map $H$.
The hash map $H$ maps the normalized coefficients $\norm(\ithRow)$ of $\ithRow$ for each row $i$ to the minimum index of its non-redundant parallel rows.
Each normalized coefficient of $\norm(\ithRow)$ is the original coefficient divided by the absolute value of the nonnegative greatest common divisor among all non-zero coefficients of $\ithRow$ (more formally, $\size[\gcd(N_z(A_{i\bullet})]$).
It is easily verified that the $i$-th and $k$-th constraints are parallel iff $\norm(\ithRow) = \norm(\kthRow)$ or $\norm(\ithRow) = -\norm(\kthRow)$. 
If $H[\norm(\ithRow)]$ does not exist, we directly let $H[\norm(\ithRow)] = i$.
Suppose that $H[\norm(\ithRow)] = k$. 
If $b_i < b_k$, then the $k$-th constraint is redundant.
Thus, we update $H[\norm(\ithRow)] = i$ and remove the $k$-th constraint from the system.
Otherwise, the $i$-th constraint is redundant and removed.
In the case where $H[-\norm(\ithRow)] = k$ and $b_i < -b_k$, we directly returns an inconsistent system since the $i$-th and $k$-th constraints are conflict.

%
%
%
%

\looseness=-1
The fourth method, namely \textit{removing subset rows}, aims to identify the subset relationships among constraints. 
We say that the $i$-th constraint is a \textit{subset} of the $k$-th constraint iff $A_{iN_z(\ithRow)} = A_{kN_z(\ithRow)}$ and $N' \neq \emptyset$ where $N' = N_z(\kthRow) \setminus N_z(\ithRow)$.
The method finds every subset constraint $i$ for each constraint $k$ in $\sysLinCons$.
If $\inf (A_{k N'}) \geq b_k - b_i$, the $i$-th constraint is dominated by the $k$-th one and thus removed.
Similarly, if $\sup(A_{k N'}) \leq b_k - b_i$, we remove the $k$-th constraint.
This method can be done in time $O(\size[M]^2 \size[N])$. 



\subsubsection{Putting It All Together}
\begin{algorithm}[t!]
	\caption{{\tt Simplify}}
	\label{alg:simplify}
	\KwIn{$\sysLinCons$: A system of integer linear constraints;}
	\KwOut{$\sysLinCons$: A simplified system.}
	
	\Repeat {$\sysLinCons = \sysLinCons'$} {
		$\sysLinCons' \assign \sysLinCons$
		
		$\sysLinCons \assign$ {\tt RemoveVariable}($\sysLinCons$)
		
		$\sysLinCons \assign$ {\tt StrengthenBound}($\sysLinCons$)
		
		\lIf { $\sysLinCons = \conflict$ } {
			\Return $\sysLinCons$
		}
		
		$\sysLinCons \assign$ {\tt RemoveIndividualRow}$(\sysLinCons)$
		
		\lIf { $\sysLinCons = \conflict$ or $M = \emptyset$} {
			\Return $\sysLinCons$
		}
		
	}
	
	$\sysLinCons \assign$ {\tt StrengthenCoefficient}$(\sysLinCons)$
	
	$\sysLinCons \assign$ {\tt RemoveIndividualRowLP}$(\sysLinCons)$
	
	\lIf { $\sysLinCons = \conflict$ or $M = \emptyset$} {
		\Return $\sysLinCons$
	}
	
	$\sysLinCons \assign$ {\tt RemoveParallelRow}$(\sysLinCons)$
	
	\lIf { $\sysLinCons = \conflict$ } {
		\Return $\sysLinCons$
	}
	
	$\sysLinCons \assign$ {\tt RemoveSubsetRow}$(\sysLinCons)$
	
	\Return $\sysLinCons$
	
\end{algorithm}

\looseness=-1
The whole framework of the above simplification techniques is shown in Algorithm \ref{alg:simplify}.
Algorithm \ref{alg:simplify} first repeatedly performs the three techniques: removing variables, strengthening bounds and removing individual rows until they cannot simplify $\sysLinCons$ anymore.
Then, the other four techniques: strengthening coefficients, removing individual rows with linear programming, removing parallel rows and removing subset rows are sequentially executed.
The reasons that the latter four techniques are not repeatedly executed are (1) the complexity of removing individual rows with linear programming is relatively high since $L$ is large for a system of linear constraints; (2) the remaining three techniques show no observable difference after repeated executions.
%
In addition, if one technique may discover a valid (or inconsistent) system of linear constraints, then Algorithm \ref{alg:simplify} directly returns a valid (or inconsistent) system after performing this technique.


\looseness=-1
\subsubsection{Discussions}
Numerous simplification techniques for mixed integer programming are proposed, which is surveyed in \cite{AchBG2020}.
However, not all of these techniques are applicable to MCILC.
Mixed integer programming focus on finding a solution that satisfies all linear constraints and that maximizes an objective function where variables may be real-valued.
In contrast, MCILC aims to count the number of solutions to all integer linear constraints.
We hereafter illustrate the techniques not adopted in this paper:
(1) Some techniques are applicable only when variables meet some conditions.
For example, probing \cite{Sav1994} and clique merging are suitable only for constraints with binary variables.
Strengthening semi-continuous and semi-integer bounds requires variables have nonnegative lower bounds.
(2) Removing parallel columns merge two variables into one, which affects the number of integer solutions.
For the same reason, neither removing dominated columns \cite{DarSM2008} nor aggregating pairs of symmetric variables is suitable to MCILC.
(3) Some techniques (\eg, cancelling nonzero coefficients \cite{ChaM1993} and extension of dual fixing) are tailored for equality constraints while our instances contains only inequality constraints.
In addition, \citet{GleHG2023} developed a C++ library includes many simplification techniques for mixed integer programming, but it do not support the three techniques: removing individual rows (with linear programming) and removing subset rows used in this paper.


\looseness=-1
In constraint programming, several high-level and solver-independent modeling systems were developed, such as MiniZinc \cite{NetSB2007} and Savile Row \cite{NigAG2017}.
These systems automatically translate a constraint model defined in their languages to the input format of various constraint, linear programming or SAT solvers.
During the translation process, the original model will be reformulated as a simple model by a collection of techniques (\eg, aggregation \cite{FriMW2002}, common sub-expression elimination \cite{Coc1970}, domain filtering \cite{BesCDL2011}, variable unification).
However, the simplification techniques employed in this paper are tailored to integer linear constraint as opposed to the richer constraint modeling language.
Notably, domain filtering and variable unification are similar techniques in constraint programming to strengthening bounds and removing variables, respectively.


%
%
%
%

\section{Empirical Evaluation}
\looseness=-1

\subsection{Implementation and Experimental Setup}
\looseness=-1
We implement our exact inter counter {\tt EDPLLSim} in C++.
Since the number of solutions and some coefficients are greater than $2^{64}$ in some instances, we utilize the GNU multiple precision arithmetic library\footnote{\url{https://www.gmplib.org/}} for handling arbitrary precision integers.
In addition, we employ the GNU linear programming kit\footnote{\url{https://www.gnu.org/software/glpk/}} as the underlying linear programming solver used in removing individual rows with linear programming.

\looseness=-1
We adopt the benchmarks used in \cite{GeB2021} that consists of the following two categories:
\begin{itemize}	
	\looseness=-1
	\item \textbf{Random benchmarks} contains 2840 instances, each characterized by three parameters $n$, $m$ and $l$ where $n \in [5, 20]$ denotes the number of variables, 
	$m \in [1, n]$ represents the number of linear constraints, and $l \in [1, n]$ indicates the maximum number of nonzero coefficients for each constraint.
	The domain of each variable is $[-8, 7]$.
	
	\looseness=-1
	\item \textbf{Application benchmarks} includes 4131 instances in total. 
	Among these, 3987 instances are generated by unwinding loops in 8 programs (cubature, gjk, http-parser, muFFT, SimpleXML, tcas, timeout and ShellSort) ranging from 400 to 7700 lines of source code.
	These instances are produced by a bug finding tool based on symbolic execution called CAnalyze \cite{XuZXW2014}.
	An additional 150 instances are generated based on the structure of simple temporal networks \cite{HuaLO2018}.
    The boundaries of the number of variables, linear constraints, the maximum number of nonzero coefficients are $[0, 28]$, $[0, 272]$, and $[0, 4]$ respectively.
	The domain of each variable is $[-32, 31]$.
\end{itemize}

\looseness=-1
We compare our approach with 7 state-of-the-art exact model counter: 1 counter for finite-dmmain constraint networks (\texttt{Cn2mddg} \cite{KorLM2015}), 2 counters for integer linear constraints (\texttt{IntCount} \cite{GeB2021} and \texttt{Barvinok} \cite{VerSB2007}) and 4 counters for propositional logic (\texttt{Cachet} \cite{SangBB2004}, \texttt{D4}\footnote{\texttt{D4} has two versions: v1.0 (\url{https://github.com/crillab/d4}) and v2.0 (\url{https://github.com/crillab/d4v2}). Since the former outperforms the latter on our task, we only choose v1.0 for comparison.} \cite{LagM2017b}, \texttt{SharpSAT-TD}\footnote{We do not present the results of \texttt{SharpSAT} \cite{Thu2006} since \texttt{SharpSAT-TD} is an improvement of \texttt{SharpSAT} with better heuristics based on tree decomposition and performs better than \texttt{SharpSAT} on the benchmarks used in this paper.} \cite{KorJ2021} and \texttt{ExactMC} \cite{LaiMY2021}).
Original application benchmarks are SMT(LA) formulas, and thus we translate them into linear constraints first.
We find 1089 instances become inconsistent(i.e. the number of variables and constraints become 0) after translation, because there are obvious conflicts in them.
In addition, we apply the tool \texttt{Boolector} \cite{NiePW2018} to translate integer linear constraints to propositional formula in conjunctive normal form for propositional model counters.
This translation typically introduces numerous auxiliary variables in the resulting propositional formula.
Some existing propositional preprocessing techniques are able to eliminate these introduced intermediate variables. 
To make a more comprehensive comparison, we further evaluate the 4 propositional model counters enhanced with the state-of-the-art preprocessing tool \texttt{Arjun} \cite{SoosM2024}.
Both translation and preprocessing times for propositional model counters are included from all reported running times.

\looseness=-1
All experiments are conducted on a machine with AMD Ryzen 9 7950X @ 4.5GHz CPU and 64 GB RAM under Ubuntu 22.04,
The time limit for each benchmark is 3600 seconds and the memory limit is 10GB.

\begin{table}[!t]
	\centering
	\fontsize{7}{10}
	\selectfont
	\setlength{\tabcolsep}{3pt}
	\caption{The numbers of benchmarks completed in total, uniquely and fastest and by every exact counter, and the average runtime on application benchmarks.}
	\vspace*{-3mm}
	\label{tab:comparison1}
	\begin{tabular}{|l|ccccccc|}
		\hline
		\multicolumn{1}{|c|}{\multirow{3}{*}{Tools}} &
		\multicolumn{7}{c|}{Benchmarks} \\ \cline{2-8} 
		\multicolumn{1}{|c|}{} &
		\multicolumn{3}{c|}{Random} &
		\multicolumn{4}{c|}{Application} \\ \cline{2-8} 
		\multicolumn{1}{|c|}{} &
		\multicolumn{1}{c|}{Total} &
		\multicolumn{1}{c|}{Unique} &
		\multicolumn{1}{c|}{Fastest} &
		\multicolumn{1}{c|}{Total} &
		\multicolumn{1}{c|}{Unique} &
		\multicolumn{1}{c|}{Fastest} &
		Time \\ \hline
		Barvinok &
		\multicolumn{1}{c|}{1,105} &
		\multicolumn{1}{c|}{0} &
		\multicolumn{1}{c|}{209} &
		\multicolumn{1}{c|}{4,129} &
		\multicolumn{1}{c|}{0} &
		\multicolumn{1}{c|}{\textbf{1,734}} &
		2.98 \\ \hline
		Cachet &
		\multicolumn{1}{c|}{921} &
		\multicolumn{1}{c|}{0} &
		\multicolumn{1}{c|}{0} &
		\multicolumn{1}{c|}{4,096} &
		\multicolumn{1}{c|}{0} &
		\multicolumn{1}{c|}{6} &
		5.34 \\ \hline
		Cachet+Arjun &
		\multicolumn{1}{c|}{966} &
		\multicolumn{1}{c|}{0} &
		\multicolumn{1}{c|}{0} &
		\multicolumn{1}{c|}{4,111} &
		\multicolumn{1}{c|}{0} &
		\multicolumn{1}{c|}{211} &
		8.14 \\ \hline
		Cn2mddg &
		\multicolumn{1}{c|}{562} &
		\multicolumn{1}{c|}{58} &
		\multicolumn{1}{c|}{62} &
		\multicolumn{1}{c|}{-} &
		\multicolumn{1}{c|}{-} &
		\multicolumn{1}{c|}{-} &
		- \\ \hline
		D4 &
		\multicolumn{1}{c|}{1,152} &
		\multicolumn{1}{c|}{0} &
		\multicolumn{1}{c|}{0} &
		\multicolumn{1}{c|}{4,101} &
		\multicolumn{1}{c|}{0} &
		\multicolumn{1}{c|}{8} &
		6.07 \\ \hline
		D4+Arjun &
		\multicolumn{1}{c|}{1,263} &
		\multicolumn{1}{c|}{0} &
		\multicolumn{1}{c|}{0} &
		\multicolumn{1}{c|}{4,118} &
		\multicolumn{1}{c|}{0} &
		\multicolumn{1}{c|}{23} &
		6.75 \\ \hline
		ExactMC &
		\multicolumn{1}{c|}{1,321} &
		\multicolumn{1}{c|}{0} &
		\multicolumn{1}{c|}{0} &
		\multicolumn{1}{c|}{4,115} &
		\multicolumn{1}{c|}{0} &
		\multicolumn{1}{c|}{0} &
		9.18 \\ \hline
		ExactMC+Arjun &
		\multicolumn{1}{c|}{1,335} &
		\multicolumn{1}{c|}{0} &
		\multicolumn{1}{c|}{3} &
		\multicolumn{1}{c|}{4,121} &
		\multicolumn{1}{c|}{0} &
		\multicolumn{1}{c|}{628} &
		10.84 \\ \hline
		IntCount &
		\multicolumn{1}{c|}{1,324} &
		\multicolumn{1}{c|}{0} &
		\multicolumn{1}{c|}{0} &
		\multicolumn{1}{c|}{4,130} &
		\multicolumn{1}{c|}{0} &
		\multicolumn{1}{c|}{987} &
		1.59 \\ \hline
		SharpSAT-TD &
		\multicolumn{1}{c|}{1,369} &
		\multicolumn{1}{c|}{63} &
		\multicolumn{1}{c|}{63} &
		\multicolumn{1}{c|}{4,092} &
		\multicolumn{1}{c|}{0} &
		\multicolumn{1}{c|}{0} &
		38.85 \\ \hline
		SharpSAT-TD+Arjun &
		\multicolumn{1}{c|}{1,470} &
		\multicolumn{1}{c|}{2} &
		\multicolumn{1}{c|}{14} &
		\multicolumn{1}{c|}{2,604} &
		\multicolumn{1}{c|}{0} &
		\multicolumn{1}{c|}{0} &
		11.99 \\ \hline
		\hline
		EDPLLSim &
		\multicolumn{1}{c|}{\textbf{1,718}} &
		\multicolumn{1}{c|}{\textbf{170}} &
		\multicolumn{1}{c|}{\textbf{1,491}} &
		\multicolumn{1}{c|}{\textbf{4,131}} &
		\multicolumn{1}{c|}{\textbf{1}} &
		\multicolumn{1}{c|}{533} &
		\textbf{0.08} \\ \hline
	\end{tabular}
	
\end{table}

\subsection{Comparative Analysis}

\begin{figure*}[tp]
	\centering
	\begin{subfigure}[b]{0.45\textwidth}
		\centering
		\includegraphics[width=0.9\linewidth]{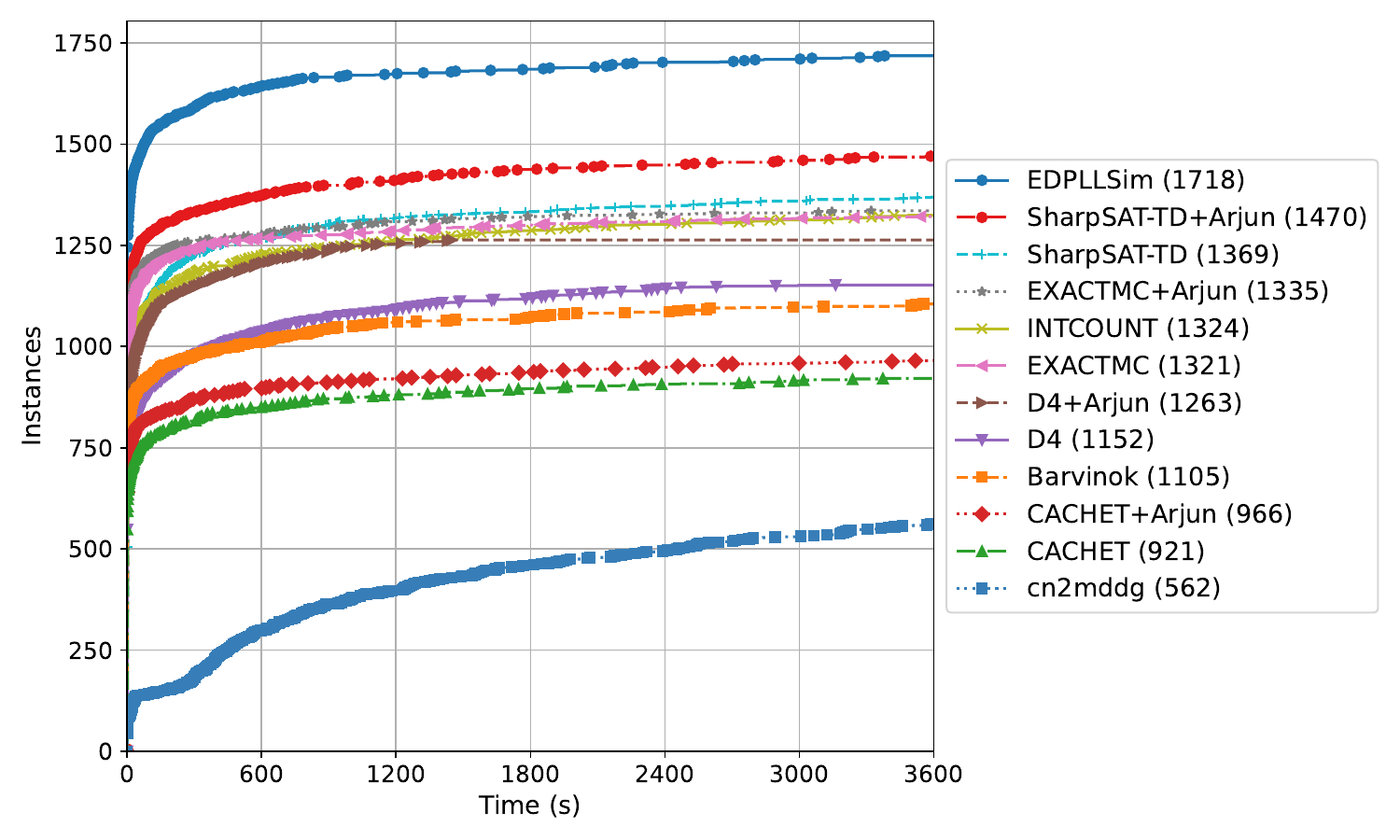}
		\caption{Random benchmarks}
		\label{fig:random}		
	\end{subfigure}
	\hfill
	\begin{subfigure}[b]{0.45\textwidth}
		\centering
		\includegraphics[width=0.9\linewidth]{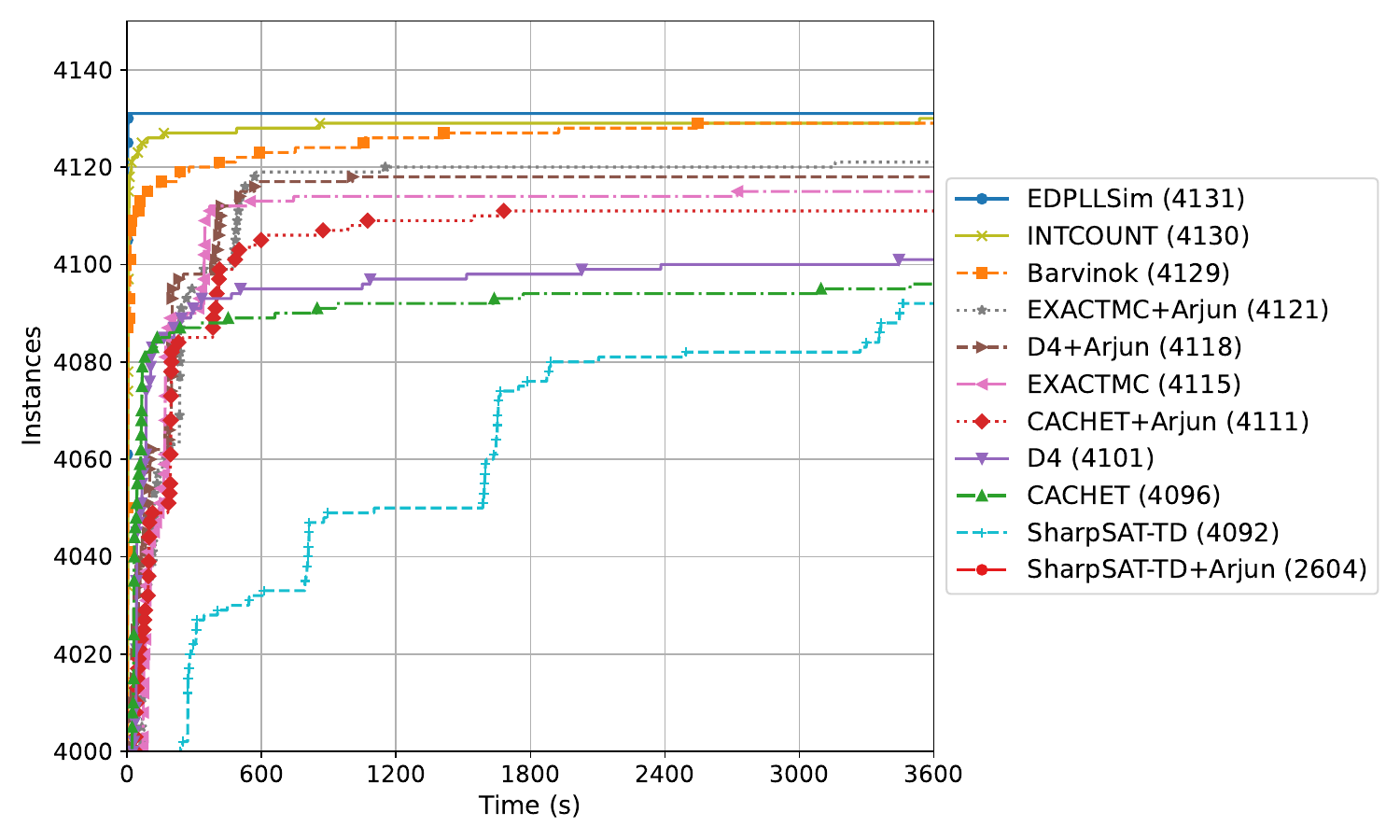}
		\caption{Application benchmarks}
	\end{subfigure}
	\caption{The number of instances solved by different model counters in a certain amount of time.}
	\vspace*{-3mm}
	\label{fig:running times}
\end{figure*}

\begin{figure*}[!h]
	\begin{subfigure}[b]{0.3\textwidth}
		\centering
		\includegraphics[width=0.95\linewidth]{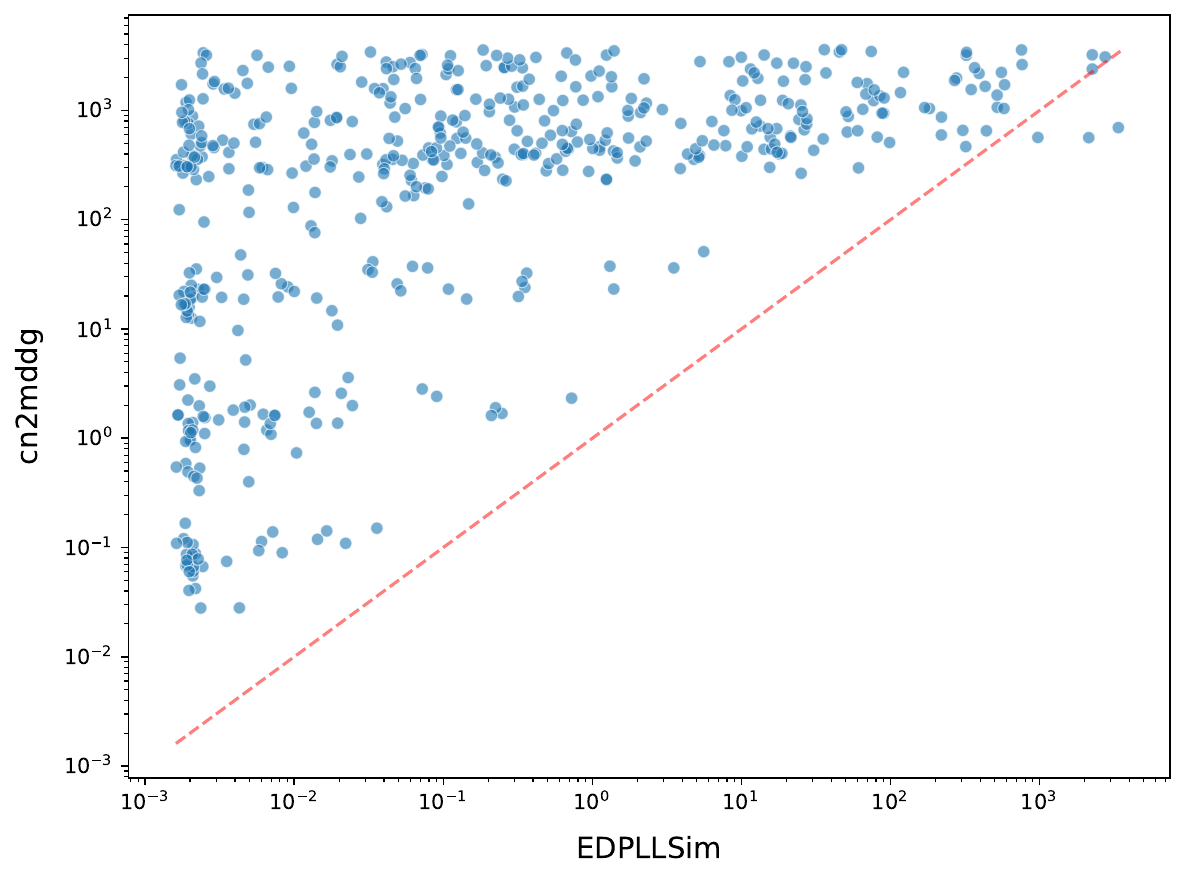}
		\caption{EDPLLSim vs Cn2mddg}
	\end{subfigure}
	\hfill
	\begin{subfigure}[b]{0.3\textwidth}
		\centering
		\includegraphics[width=0.95\linewidth]{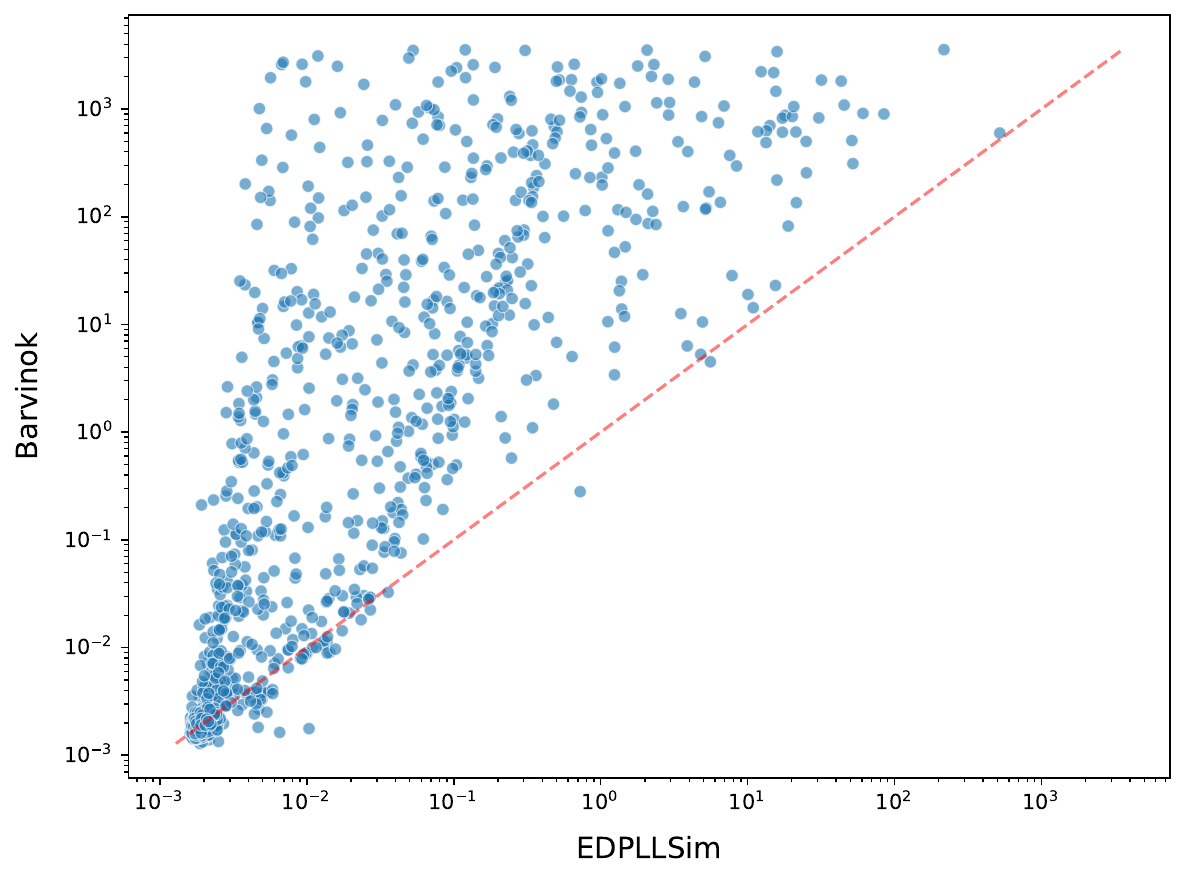}
		\caption{EDPLLSim vs Barvinok}		
	\end{subfigure}
	\hfill
	\begin{subfigure}[b]{0.3\textwidth}
		\centering
		\includegraphics[width=0.95\linewidth]{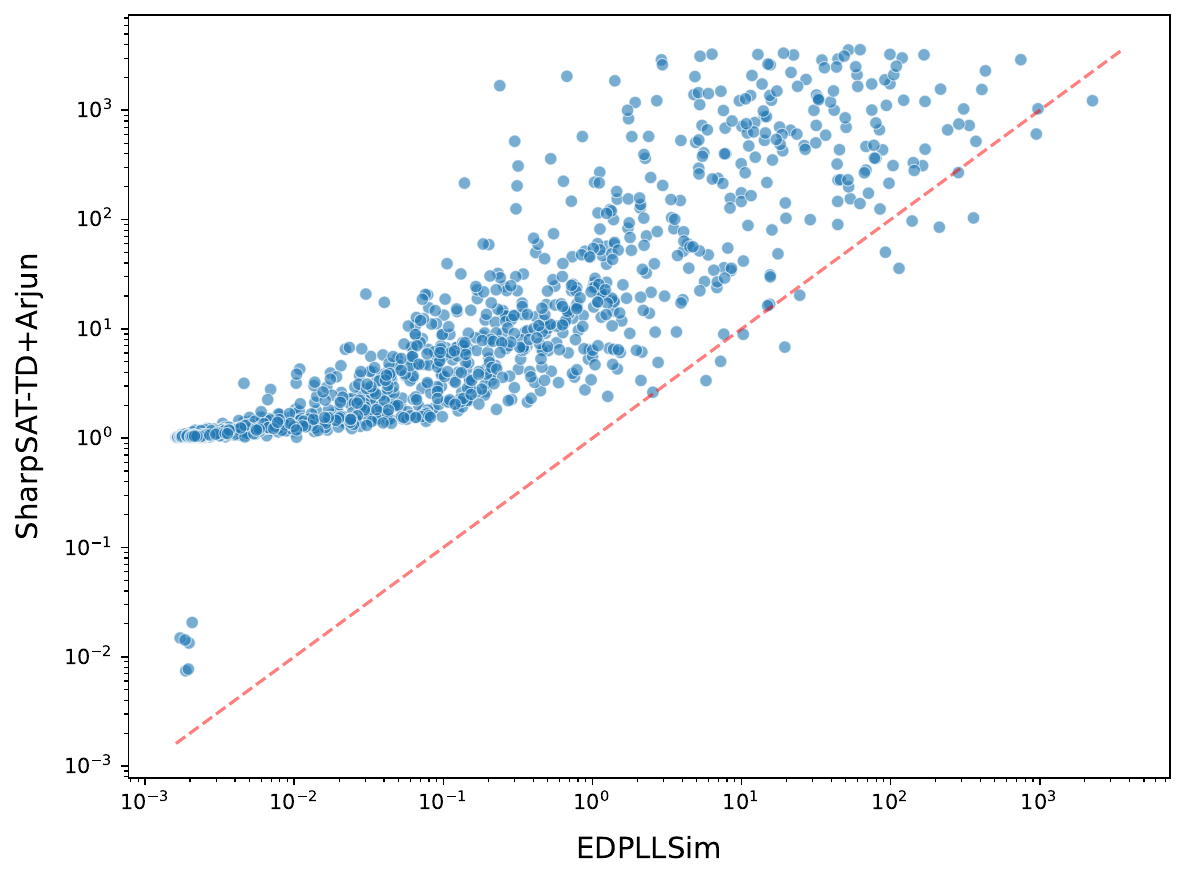}
		\caption{EDPLLSim vs SharpSAT-TD+Arjun}
	\end{subfigure}
	\caption{Comparison on the commonly solved benchmarks}
	\vspace*{-3mm}
	\label{fig:perfComp}
\end{figure*}


\looseness=-1
Figure \ref{fig:running times} illustrates the number of instances solved by each model counter within a fixed time limit.
Table \ref{tab:comparison1} provides a detailed comparison of their performance.
The columns ``Total", ``Unique", and ``Fastest" indicate the total number of solved instances, of instances uniquely solved by the corresponding counter, and of instances for which it achieved the fastest solving time, respectively.
The best results are highlighted in bold.



\looseness=-1
\subsubsection{Random Benchmarks}
We can make several observation from Figure \ref{fig:running times}(a) and Table \ref{tab:comparison1}.
(1) Among all exact model counters, \texttt{EDPLLSim} solves the most instances on random benchmarks.
Specifically, it is able to solve 1718 instances, which is 248 more than \texttt{SharpSAT-TD} with \texttt{Arjun} that ranks second.
(2) With the limit of 78 seconds per instance, \texttt{EDPLLSim} is already able to solve above 1500 instances, which is more than the total number of instances solved by \texttt{SharpSAT-TD} with \texttt{Arjun} within 3600 seconds.
(3) It is worth noting that \texttt{EDPLLSim} solves 1491 instances faster than other counters, accounting for 86.8\% of the solved instances.
(4) \texttt{EDPLLSim} also solves 168 instances uniquely (\ie, these instances cannot be solved by other counters), which ranks fist as well.
In summary, \texttt{EDPLLSim} outperforms all exact counters remarkably on random benchmarks.

\looseness=-1
Nevertheless, there are specific instances on which other counters exhibit better performance.
\texttt{Cn2mddg} uniquely solves 58 instances and shows a runtime advantage on 62 instances.
These instances involve between 2 and 9 constraints, indicating that \texttt{Cn2mddg} is well-suited for MCILC instances with a small number of constraints. 
Figure 2(a) displays the runtime comparison on common benchmarks solved by both \texttt{Cn2mddg} and \texttt{EDPLLSim}.
A total of 503 instances are solved by both approaches, among which EDPLLSim is faster on 500 instances.
\texttt{Cn2mddg} shares the same exhaustive DPLL architecture and decomposition way as \texttt{EDPLLSim}, but it solves only 562 instances of which has at most 9 constraints.
This is because \texttt{Cn2mddg} has to compile the constraint network into a MDDG, and thus requires substantial space overhead.
Moreover, it takes no advantage of specific simplification techniques for MCILC.

\looseness=-1
It can be observed that \texttt{Barvinok} runs the fastest on 209 instances.
These instances have at most 7 nonzero coefficients in each constraint with an average of only 1.9.
%
%
This means that \texttt{Barvinok} performs efficiently when facing simple MCILC problems. 
However, it solves 613 fewer instances than \texttt{EDPLLSim}.
As shown in Figure 2(b), besides the above 209 instances, EDPLLSim is faster on 896 out of the 1105 instances commonly solved by \texttt{Barvinok} and \texttt{EDPLLSim}.
\texttt{IntCount}, which integrates a preprocessing stage that eliminates redundant rows and columns, solves 219 more instances than directly applying Barvinok's algorithm. 
However, this is still 394 fewer than the number of instances solved by \texttt{EDPLLSim}.

\looseness=-1
Finally, we analyze the performance of propositional model counting-based approaches.
It can be observed that the propositional preprocessor \texttt{Arjun} significantly improve the number of instances solved by propositional model counters, albeit with some additional time overhead.
For example, \texttt{D4} is able to solve 111 more instances with \texttt{Arjun}.
A total of 65 instances are uniquely solved by the propositional model counters \texttt{SharpSAT-TD} with or without \texttt{Arjun}.
Notably, \texttt{SharpSAT-TD} with \texttt{Arjun} ranks as the second-best approach for MCILC.
However, in terms of instances coverage, our approach \texttt{EDPLLSim} outperforms all propositional model counting-based approaches.
In particular, \texttt{EDPLLSim} outperforms \texttt{SharpSAT-TD} with \texttt{Arjun} on 1455 of the 1468 commonly solved instances.

\looseness=-1

\subsubsection{Application Benchmarks} 
\looseness=-1
All model counters except \texttt{Cn2mddg} and \texttt{SharpSAT-TD} with \texttt{Arjun} are able to solve the majority of instances in application benchmarks.
Because \texttt{Cn2mddg} cannot handle arbitrary precision numbers so it solves no instances.
\texttt{SharpSAT-TD} with \texttt{Arjun} reports error (i.e. program interrupted) on 1520 instances which we exclude in the experimental result.
Table \ref{tab:comparison1} summarizes the detailed performance of counters. 
In addition, we show the average runtime of these counters in column ``Time" on application benchmarks. 

\looseness=-1
As observed in Table \ref{tab:comparison1}, \texttt{EDPLLSim} is the only one that can solve all instances (in total 4131).
The instance solved by \texttt{EDPLLSim} uniquely is STN\_T1\_17 with 17 variables and 272 linear constraints, 
\texttt{EDPLLSim} counts its models within 0.07 seconds but no other counter successfully solves it in 3600 seconds.
Furthermore, it can solve all application instances in a short time.
The average runtime of \texttt{EDPLLSim} over 4131 instances is only 0.08 seconds yields a speedup of 20$\times$ over \texttt{IntCount} which is the second fastest counter.
These experimental results show that \texttt{EDPLLSim} outperforms all counters on application benchmarks in general. 

\looseness=-1
Other MCILC counters, \texttt{Barvinok} and \texttt{IntCount}, achieves better runtime efficiency on certain instances.
To be specific, \texttt{Barvinok} achieves the fastest solving time on 1734 instances with up to 19 variables and at most 110 constraints.
Meanwhile, \texttt{IntCount} performs best on another 987 instances with at most 23 variables and 69 constraints.

\looseness=-1
Although propositional model counters together (\texttt{Cachet}, \texttt{D4} and \texttt{ExactMC} with (or without) \texttt{Arjun}) achieve the fastest solving time on a total of 876 instances, these instances are inconsistent and trivial since their inconsistency can be easily checked. 
Besides trivial instances, propositional model counters do not perform as well as MCILC counters. 
For each instance, the best propositional model counter \texttt{Cachet} requires 5.34 seconds on average while \texttt{EDPLLSim} uses only 0.08 seconds.
These results clearly demonstrate the superiority of EDPLLSim over propositional model counters on application benchmarks.

\section{Conclusions}
\looseness=-1
The exhaustive DPLL architecture, in conjunction with connected component-based decomposition constitutes an effective framework for propositional model counting and serves as the core algorithmic foundation of many state-of-the-art propositional exact model counters.
In this paper, in order to develop an effective and exact model counter for integer linear constraints, we adopt this architecture as the backbone of our model counter.
In addition, to improve the efficiency, we integrate some powerful simplification techniques tailored for integer linear constraints into our model counter.
We have demonstrated the superior performance of our model counter \texttt{EDPLLSim} by evaluating it on both random and application benchmarks for MCILC.

%
%
%
%
%
%


\bibliography{AAAI-2026-1}

\end{document}